\documentclass[10pt, a4paper]{article}

\usepackage{times}
\usepackage{url}
\usepackage{latexsym}
\usepackage{multirow}
\usepackage{slashbox}
\usepackage{adjustbox}
\usepackage{subcaption}
\usepackage{graphicx}
\usepackage{lrec-coling2024} 

\title{Forecasting Four Business Cycle Phases Using Machine Learning: A Case Study of US and EuroZone}

\name{Elvys Linhares Pontes, Mohamed Benjannet, and Raymond Yung} 

\address{Trading Central Labs, Trading Central, Paris, France\\
         \{elvys.linharespontes,mohamed.benjannet\}@tradingcentral.com\\}

\abstract{
Understanding the business cycle is crucial for building economic stability, guiding business planning, and informing investment decisions. The business cycle refers to the recurring pattern of expansion and contraction in economic activity over time.  Economic analysis is inherently complex, incorporating a myriad of factors (such as macroeconomic indicators, political decisions). This complexity makes it challenging to fully account for all variables when determining the current state of the economy and predicting its future trajectory in the upcoming months. The objective of this study is to investigate the capacity of machine learning models in automatically analyzing the state of the economic, with the goal of forecasting business phases (expansion, slowdown, recession and recovery) in the United States and the EuroZone.  We compared three different machine learning approaches to classify the phases of the business cycle, and among them, the Multinomial Logistic Regression (MLR) achieved the best results. Specifically, MLR got the best results by achieving the accuracy of 65.25\% (Top1) and 84.74\% (Top2) for the EuroZone and 75\% (Top1) and 92.14\% (Top2) for the United States. These results demonstrate the potential of machine learning techniques to predict business cycles accurately, which can aid in making informed decisions in the fields of economics and finance.
 \\ \newline  }

\begin{document}

\maketitleabstract

\section{Introduction}

The economy is a multifaceted and dynamic system characterized by complexity, uncertainty, and interdependence. Understanding the economy requires delving into a complex web of interconnected systems, behaviors, and variables that influence the production, distribution, and consumption of goods and services within a society or across the globe.

Business cycles represent fluctuations in the overall economic activity of a country. These cycles involve periods of expansion, where economic activities across various sectors simultaneously grow, followed by periods of contraction, where economic activities similarly decline\footnote{\url{https://www.investopedia.com/terms/b/businesscycle.asp}}.
The study and prediction of the business cycle is crucial for promoting economic stability, guiding business planning, and informing investment decisions. By understanding the various phases of the business cycle, policymakers can design effective policies to stabilize the economy during downturns. Similarly, investors and businesses can use this knowledge to make decisions about where to allocate capital and how to adjust their strategies.

A business cycle is characterized by fluctuations in Gross Domestic Product (GDP) around its long-term natural growth rate, elucidating the alternating phases of expansion and contraction in economic activity that an economy undergoes over time. The delineation of these phases in the business cycle framework lacks global consensus, with varying perspectives proposing two, four\footnote{\url{https://www.indeed.com/career-advice/career-development/the-4-phases-of-the-business-cycle}}, or six phases\footnote{\url{https://corporatefinanceinstitute.com/resources/economics/business-cycle/}}. In this study, we have chosen to adopt a business cycle model consisting of four phases: expansion, slowdown, recession, and recovery.
Expansion is characterised by an economic growth and an increasing in output, employment, and income. Slowdown is the phase following expansion where economic activity start to decline. Recession is characterised by declining economic activity, decreasing output, employment, and income.
Finally, recovery is the end of recession phase and the beginning of economic recovery with increasing output, employment, and income.

Predicting the business cycle is a challenging task because business cycles themselves are essentially theoretical models that attempt to explain the variation in business data. 
Indeed, economic analysis is a multifaceted endeavor, intricately woven with a multitude of factors that influence its trajectory. At its core, it involves the examination of macroeconomic indicators such as GDP, inflation rates, employment figures, and trade balances. However, the complexity of economic analysis extends beyond these indicators to encompass a wide array of influences, including political decisions, technological advancements, social trends, and global events. This complexity makes it challenging to fully account for all variables when determining the current state of the economy and predicting its future trajectory in the upcoming months.

In the United States of America, business cycles are defined by the National Bureau of Economic Research (NBER), and in EuroZone, they are determined by the European Central Bank (ECB). However, both institutions only identify recessions, which can be estimated using a common rule of thumb: a significant decline in GDP over two consecutive quarters \cite{an2018well}.
Earlier research in the field of dating economic cycles primarily focused on using binary dependent variable models, such as Probit models, to detect and forecast turning points in the cycle. Then in the 90s, Markov switching models were developed and became widely used for predicting economic regimes.
Today, Machine learning has proven to be a valuable resource in various fields beyond the social sciences, and it has started to be more extensively used in economics.

Given the complexity of analyzing the state of the economy and forecasting the business phase of a country (region), the goal of this paper is to investigate the potential of machine learning (ML) approaches to forecast the business cycles of the United States (US) and EuroZone (EZ), with the ability to identify four distinct phases: recession, recovery, expansion, and slowdown. We compiled an annotated dataset covering the United States and EuroZone from 1970 to 2022. Then, we compared three ML approaches and a baseline in order to identify the most effective data preprocessing and ML algorithms for the business cycle task. 
Among these approaches, MLR achieved the accuracy of 65.25\% (top 1) and 84.74\% (top 2) for EZ and 75\% (top 1) and 92.14\% (top 2) for US.
These results demonstrate the potential of machine learning techniques to predict business phases, which can help inform decision-making in the fields of economics and finance.

This paper is organized as follows: we survey the business cycle prediction in Section~\ref{sc:related_work}. Next, we explain how the datasets were built on Section~\ref{sc:dataset}. Then, we detail the machine learning approaches in Section~\ref{sc:approach}. The experiments and the results are discussed in Sections~\ref{sc:setup} and~\ref{sc:evaluation}, respectively. Lastly, we provide the conclusion and some final comments in Section~\ref{sc:conc}.

\section{Related work}
\label{sc:related_work}

Initially, research efforts concentrated on utilizing a limited number of macroeconomic indicators, notably Gross National Product (GNP), to gauge a country's economic performance. Subsequently, investigations expanded to encompass a broader spectrum of macroeconomic factors, including but not limited to growth rates, interest rates, industrial production, real manufacturing and trade sales, among others. Analyzing these macroeconomic factors necessitates distinct methodological strategies, with complexity escalating as additional variables are introduced. We have therefore focused on analyzing state-of-the-art approaches to highlight the progress made in this field and assess their effectiveness in business cycle analysis.

Most of the research in the field of predicting business cycles has utilized either logit/probit regressions or Markov switching models as their preferred models.
Logit/probit models have their roots dating back to the late 1800s \cite{cramer2002origins}, while the Markov-switching models were introduced in 1989 \cite{hamilton1989new}.

In Hamilton experiment, the modifications are considered as a first-order Markov process that is not observed directly, and the probability of transitioning from one regime to another is estimated using the available macroeconomic data. Other studies have tried to enhance the Markov-switching models by changing the prediction to monthly basis \cite{filardo1994business} or by adding dynamic factors \cite{diebold1994measuring}, \cite{chauvet1995econometric}.
In \cite{chauvet2008comparison}, the authors evaluated the performance of their Markov-switching model under real-time conditions.

Many studies have also focused on studying the logit/probit models, \citet{kauppi2008predicting} analyzed the use of the yield curve to forecast a recession 1 to 6 months ahead.
\citet{fossati2016dating} compared the performance of static and dynamic probits to Markov switching models for business cycle phases, using separate models for "small data" and "big data" factors. They found that static probit models performed well in detecting cyclical peaks and troughs sooner, but with more volatile probability outputs and false positives.
\citet{katayama2009improving} examined the use of logistic regression models to predict recessions 6 months ahead. They found that a trivariate model based on the yield curve, 3-month change in the S\&P 500, and nonfarm employment performed the best. The use of Laplace and Gumbel cumulative density functions resulted in superior predictive performance compared to more commonly used normal and logistic cumulative density functions (CDFs).

Machine learning has been proven to be a useful tool in several applications in finance, e.g. the sentiment classification of tweets and headlines in finance~\cite{Pontes2021} and the recognition of ESG concepts~\cite{linhares-pontes-etal-2022-using}.
Many studies have investigated its use for predicting economic regimes, \citet{james2019machine} used Support Vector Machines (SVM) to predict the begin and the end of recession for US economy.
\citet{kamal2021analysis} aimed to identify a supervised machine learning method to predict recessions for US economy, considering logistic regression, decision tree classifier, $k$ nearest neighbor classifier, and SVM. The findings indicate that the most effective model is the tuned support vector classifier model trained on scaled data.
In \cite{vrontos2021modeling}, authors used a comparative framework to evaluate machine learning techniques for estimating US recession probabilities. The results indicate that penalized logit regression models, $k$-nearest neighbors, and bayesian generalized linear models perform better than traditional econometric techniques like logit/probit models.
Finally, \citet{wang2022economic} studied the use of deep learning methods for predicting recessions and found out that the Bi-LSTM with Autoencoder is accurate for this task.

Taking in consideration of four possible business phases, \citet{Morik:2002} investigated the applicability of inductive logic programming (ILP) to the problem of predicting the business phases on the German business cycle data. While their ILP model for four phases got only the average accuracy of 53\%; their two phase model achieved 82\%.

By analyzing Wall Street Journal articles from 1984 to 2017, \citet{NBERw29344} proposed a text-augmented VAR model that demonstrates the significant additional contribution of news text in modeling macroeconomic dynamics and allows for the retrieval of narratives underlying business cycle fluctuations.

\citet{NBERw29344} discuss an alternative approach to understanding economic fluctuations by using narratives in business news rather than relying solely on numerical macroeconomic indicators. The authors argue that news media serves as a verbal mirror of the economy, providing information on economic events and their interpretation. The authors propose using this narrative-focused approach to enhance the interpretation of economic models, suggesting it can offer detailed insights into the drivers of economic dynamics. 

In this study, we are following the trend of using machine learning (ML) for predicting economic cycles by comparing different ML methods. Unlike most other studies in the literature, we are not only attempting to predict recession dates (binary classification), but also aiming to predict four phases of the business cycle: expansion, slowdown, recession, and recovery. We evaluated these models on two economic regions: US and EZ. Forecasting the business cycle in EZ is particularly difficult due to the need to integrate data from multiple EZ countries and the fact that the historical data available for the EZ region as an integrated economic area is relatively shorter than for the United States.

\section{Annotated dataset}
\label{sc:dataset}

As mentioned earlier, most researches in this field has primarily focused on identifying recession periods. However, in our study, we aimed to analyze the task of predicting four possible phases of the business cycle for United States and EuroZone. 

Firstly, we calculated the inflation and growth indices for each region/nation. To do this, we first identified the macroeconomic series related to growth and inflation and grouped them into those two categories, respectively. Secondly, we checked the stationarity of those time series and performed appropriate data transformations when necessary. Thirdly, as macroeconomic data are released from various sources at different times and frequencies, we transformed and standardized the raw data by Z-score calculation\footnote{\url{https://www.investopedia.com/terms/z/zscore.asp}}. In addition, we addressed the autocorrelation issue of the data by using the Newey-West Estimator in conjunction with a subsampling technique. As a final step, we applied PCA (Principal Component Analysis) \cite{Pearson1901pca} to extract the key driver of the large dataset and build the index using an expanding window with a minimum period of 5 years. The output of the PCA (the first component) is our inflation and growth indices.

Then, we created two monthly datasets (United States and EuroZone) that classify each month into a specific business cycle phase. To ensure the consistency of the annotation of business phases, we defined and annotated each phase based on the growth and inflation indices as follows:
\begin{itemize}
    \item Recession: This phase is typically characterized by negative GDP growth, lower inflation driven by excess production capacity, weak profits, and dropping real yields. 
    In our dataset, the recession phases are the same as the official recession and the technical recession declared by the official organization of each region. 
    \item Recovery: During an economic recovery, inflation, as measured by the Consumer Price Index (CPI) and general import prices, typically decreases. This is because production capacity has not yet reached its limit, and cyclical productivity is on the rise. At the same time, indicators of growth, such as GDP growth, Purchasing Managers' Index (PMIs), and retail sales, often rises. This rise is usually accompanied by a sharp rebound in profits. 
    \item Expansion: The expansion phase is characterized by positive economic growth rates with GDP growth staying above trend with rising inflation rates. 
    \item Slowdown: This phase is characterized by falling economic growth rates as productivity slumps and wage increases affect the profit margins of companies. Also inflation rates are rising or stabilizing under a loosening monetary policy as central banks are hesitant to ease until inflation peaks. 
\end{itemize}

\begin{figure}[h]
\centering
\begin{subfigure}[b]{\linewidth}
   \includegraphics[width=\linewidth]{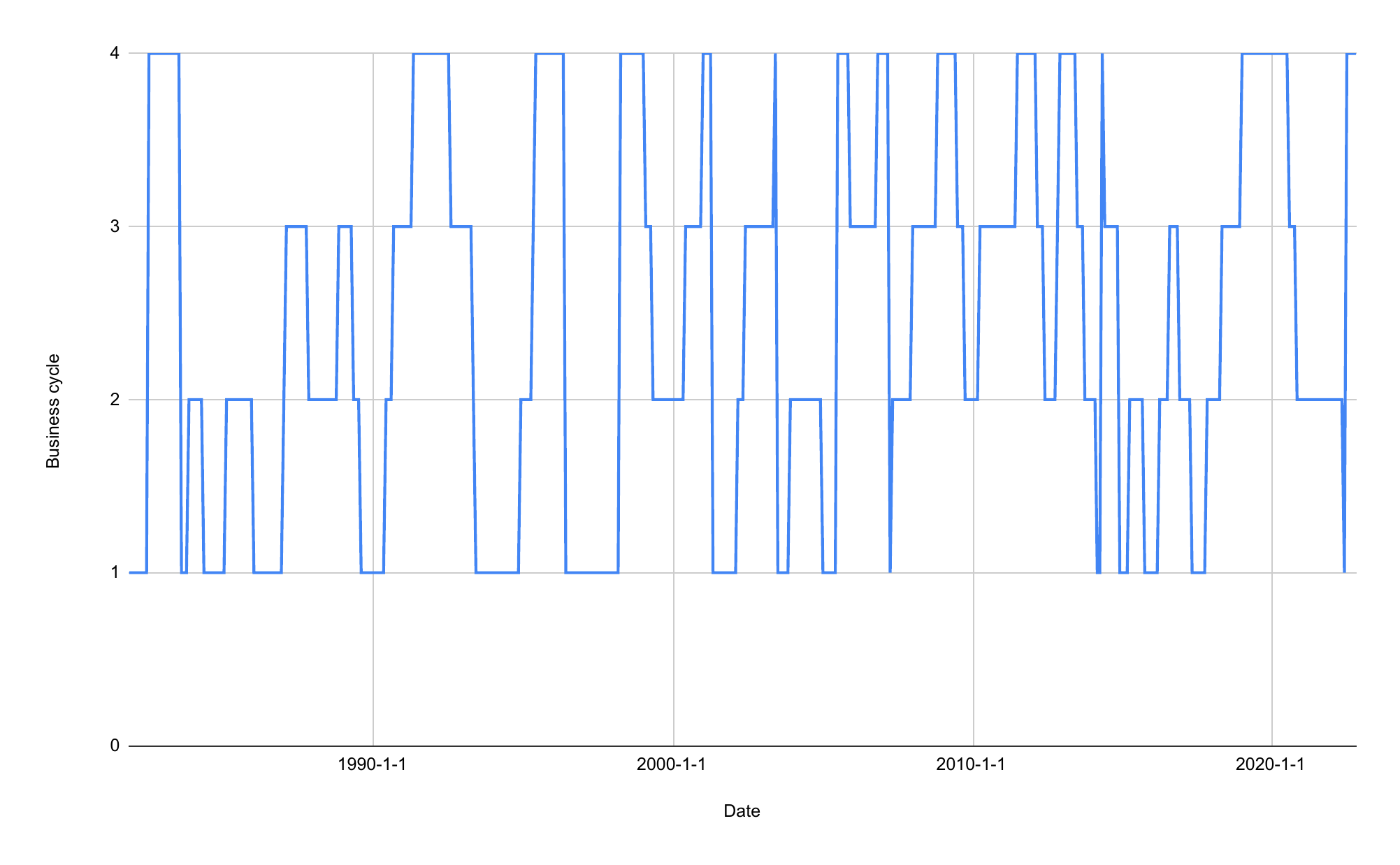}
   \caption{EuroZone.}
   \label{fig:Ng2}
\end{subfigure}
\begin{subfigure}[b]{\linewidth}
   \includegraphics[width=\linewidth]{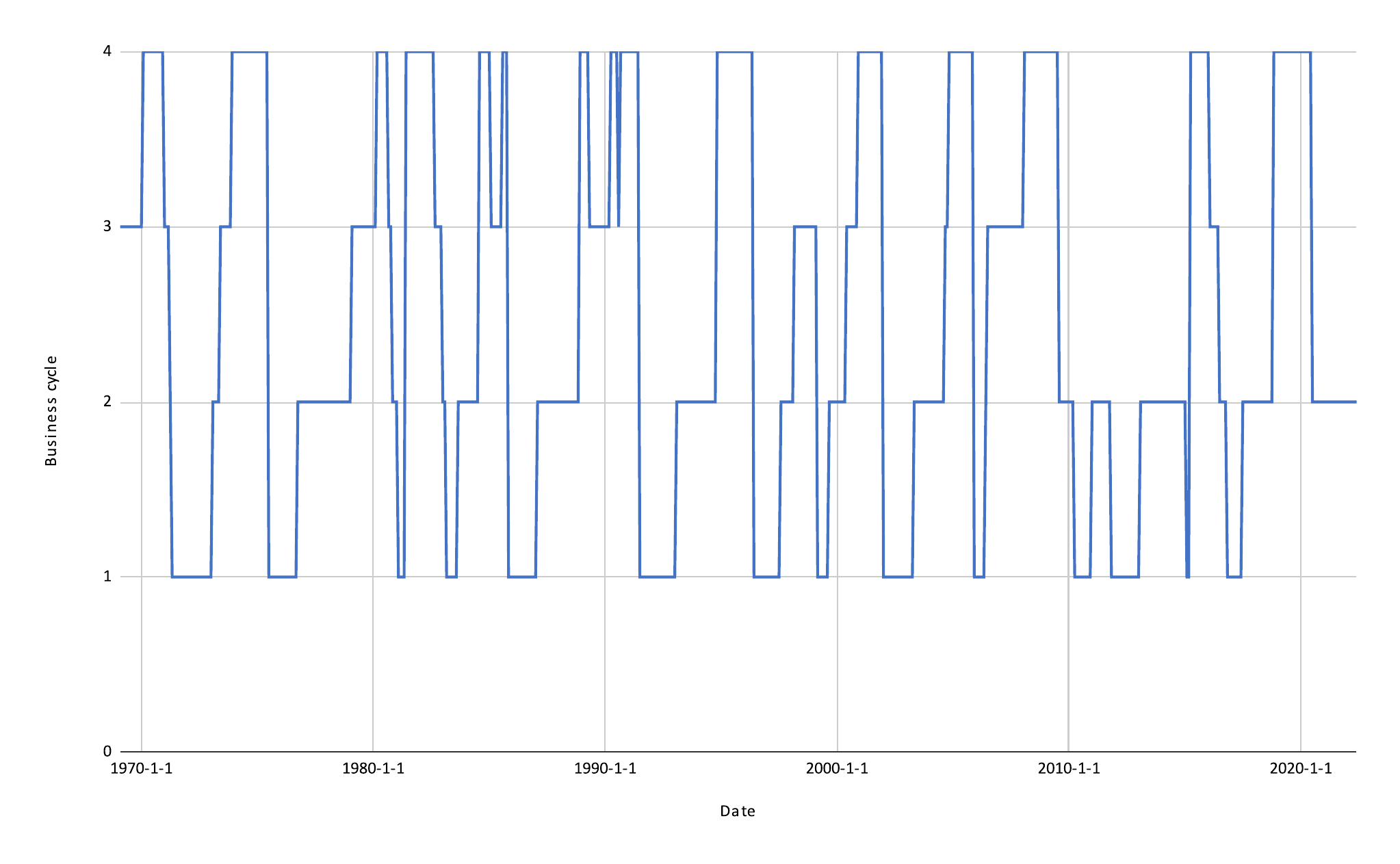}
   \caption{United States.}
   \label{fig:Ng1} 
\end{subfigure}
\caption[Two numerical solutions]{\label{img:annotated-dataset}Historic of phases of business cycle on the United States (a) and EuroZone (b). Recovery is represented by '1', expansion by '2', slowdown by '3' and recession by '4'.}
\end{figure}

Figure~\ref{img:annotated-dataset} shows our annotated dataset with the phases of the business cycle over time for the United States and the EuroZone.

\section{Machine learning approaches}
\label{sc:approach}

Our goal is to focus on obtaining a precise understanding of the business cycle by analyzing the current economic situation of a particular nation or region. By gathering and examining relevant data, we can track the evolution of the economy and provide accurate business cycle for the nation or region in question. The business cycle information enables us to gain a comprehensive view of the current state of the economy, allowing us to identify potential risks and opportunities for investments. By leveraging this information, we can help businesses, investors, and policymakers make informed decisions that are grounded in accurate, up-to-date information about the economy.

The following subsections will provide a description of the macroeconomic data (Section~\ref{ssc:macro}) utilized in our analysis, as well as the machine learning approaches (Section~\ref{ssc:ml}) employed in our evaluation.

\subsection{Macroeconomic data}
\label{ssc:macro}

The economy is a complex system that is impacted by multiple sectors. In order to gain a comprehensive understanding of the most significant factors affecting the economy, it is essential to analyze the interconnections between these sectors. To achieve this goal, we have examined several key sectors, including the central bank, commodities, and stock market.

To track the evolution of the economy across different sectors, we have considered a range of macroeconomic indicators, including the performance of specific industries such as: automobile, business services, clothing, construction, corporate, durable and non-durable goods, employees, exportation, financial activities, GDP and gross, goods and services, household, industrial, jobs, manufacturing, oil gas extraction, private service, retail trade, private production, and transportation utilities. These macroeconomic data were obtained from FRED economic data API\footnote{\url{https://fred.stlouisfed.org/}} and TradingEconomic\footnote{\url{https://tradingeconomics.com/}} for the United States and EuroStat\footnote{\url{https://ec.europa.eu/eurostat}} for the EuroZone. In addition, we have examined several indicators to gauge the health of the government, including inflation rate, treasury bills, federal funds rate, interbank rates, and market yield treasury securities.

Furthermore, to gain insights from the commodities market, we have analyzed various materials, including brent oil, coal, crude oil, gasoline, gold, industrial materials index, and iron ore. Finally, we have monitored key stock market indices, including AEX, CAC40, DAX, Dow Jones, IBEX, NASDAQ, S\&P 500, and VIX, to track performance across different sectors\footnote{We used the Yahoo Finance API to get the commodities and indices values: \url{finance.yahoo.com}.}. By analyzing these sectors and indicators, we are better equipped to understand the current state of the economy and make informed predictions about future trends.

\subsection{Machine learning methods}
\label{ssc:ml}

Machine learning approaches have a well-established reputation for their capacity to analyze and exploit the interdependencies within input data to develop high-performing models that can effectively classify this data into predefined categories. In our specific case, we aim to develop a model that analyzes macroeconomic data for a particular region and can accurately classify the current economic situation of the region into one of the phases of the business cycle. To achieve our objective, we chose three well-established machine learning approaches, namely Multinomial Logistic Regression, Support Vector Machines, and Multi-layer Perceptron, for performing supervised learning on the task of predicting the business cycle\footnote{We used the scikit learn toolkit (\url{scikit-learn.org}) for the Multinomial Logistic Regression and Support Vector Machines.}.

\subsubsection{Multinomial Logistic Regression}

Multinomial logistic regression (MLR) is a statistical model used to predict and analyze relationships between a categorical dependent variable with more than two categories and one or more independent variables that can be continuous or categorical~\cite{Böhning1992mlr}. 

This approach works by calculating the log odds (logit) of each possible outcome, where the odds of each outcome is defined as the probability of the outcome occurring divided by the probability of the outcome not occurring. The model then calculates the predicted probability of each outcome based on the log odds and the independent variables.

\subsubsection{Support vector machines}

Support vector machines (SVMs) work by finding the optimal boundary or hyperplane that separates data points into different classes or groups~\cite{Suthaharan2016}. The algorithm maximizes the margin between the hyperplane and the closest data points from each class, which helps to ensure that the boundary is as robust and generalizable as possible. 

For our task, it takes a set of labeled training data and creates a model that can predict the class of new data. This approach maps the data to a high-dimensional feature space and finds the hyperplane that best separates the data into different classes. The algorithm can handle non-linear data by using a kernel function to transform the data into a higher-dimensional space, where it may be easier to find a linear boundary.

\subsubsection{Multi-layer Perceptron}

A Multi-layer Perceptron (MLP) is a type of artificial neural network (ANN) that is composed of multiple layers of interconnected nodes. MLP is a feed-forward neural network, meaning that information flows only in one direction, from the input layer to the output layer, without any loops or feedback connections~\cite{Popescu2009mlp}. 

MLPs are commonly used for supervised learning tasks, such as classification and regression, and can be trained using back-propagation and gradient descent algorithms to minimize the loss function between the predictions and the correct targets.

\section{Experimental setup}
\label{sc:setup}

In order to evaluate the performance of ML methods, we divided the dataset into three subsets: train, validation, and test (Table~\ref{tb:split_dataset}). Due to constraints in the availability of macroeconomic data, we opted to annotate information spanning from 1969 for the United States and from 1981 for the Eurozone. This timeframe was selected based on the predominant accessibility of comprehensive macroeconomic data, which is widely available starting from these respective years. We took into consideration the frequency of phases to ensure a stable partitioning. Table~\ref{tb:statistics_dataset} shows the number of examples for each phase over time in each subset of the datasets. 

\begin{table}[h]
\begin{adjustbox}{width=\linewidth}
\begin{tabular}{lccc}
\hline
\textbf{Region} & \textbf{Train} & \textbf{Validation} & \textbf{Test} \\ \hline
EuroZone & 1981-2001 & 2002-2011 & 2012-2022 \\ 
US       & 1969-1999 & 2000-2009 & 2010-2022 \\ 
\hline
\end{tabular}
\end{adjustbox}
\caption{\label{tb:split_dataset}Time intervals considered for train, validation and test datasets.}
\end{table}

\begin{table}[h]
\begin{adjustbox}{width=\linewidth}
\begin{tabular}{llcccc}
\hline
\textbf{Region}    & \textbf{split} & \textbf{recov.} & \textbf{expans.} & \textbf{slowd.} & \textbf{reces.} \\\hline
\multirow{3}{*}{EZ} & train &    89    &    56   &    31   &    67      \\
                    & dev   &    18    &    28   &    51    &   23      \\
                    & test  &    21    &    51   &    24    &   34      \\\hline
\multirow{3}{*}{US} & train &    100   &    104  &    70    &   98      \\
                    & dev   &    25    &    24   &    27    &   44      \\
                    & test  &    37    &    77   &    8     &   27      \\\hline
\end{tabular}
\end{adjustbox}
\caption{\label{tb:statistics_dataset}This table displays the dataset statistics for each split, with each cell indicating the number of examples for each phase.}
\end{table}

Given that the majority of macroeconomic data is reported on a monthly basis, with values released on different days throughout the month, our machine learning models are designed to forecast the business phase for the upcoming month based on the macroeconomic data available for a given month.

\begin{figure}[h]
\centering
\begin{subfigure}[b]{0.45\linewidth}
   \includegraphics[width=\linewidth]{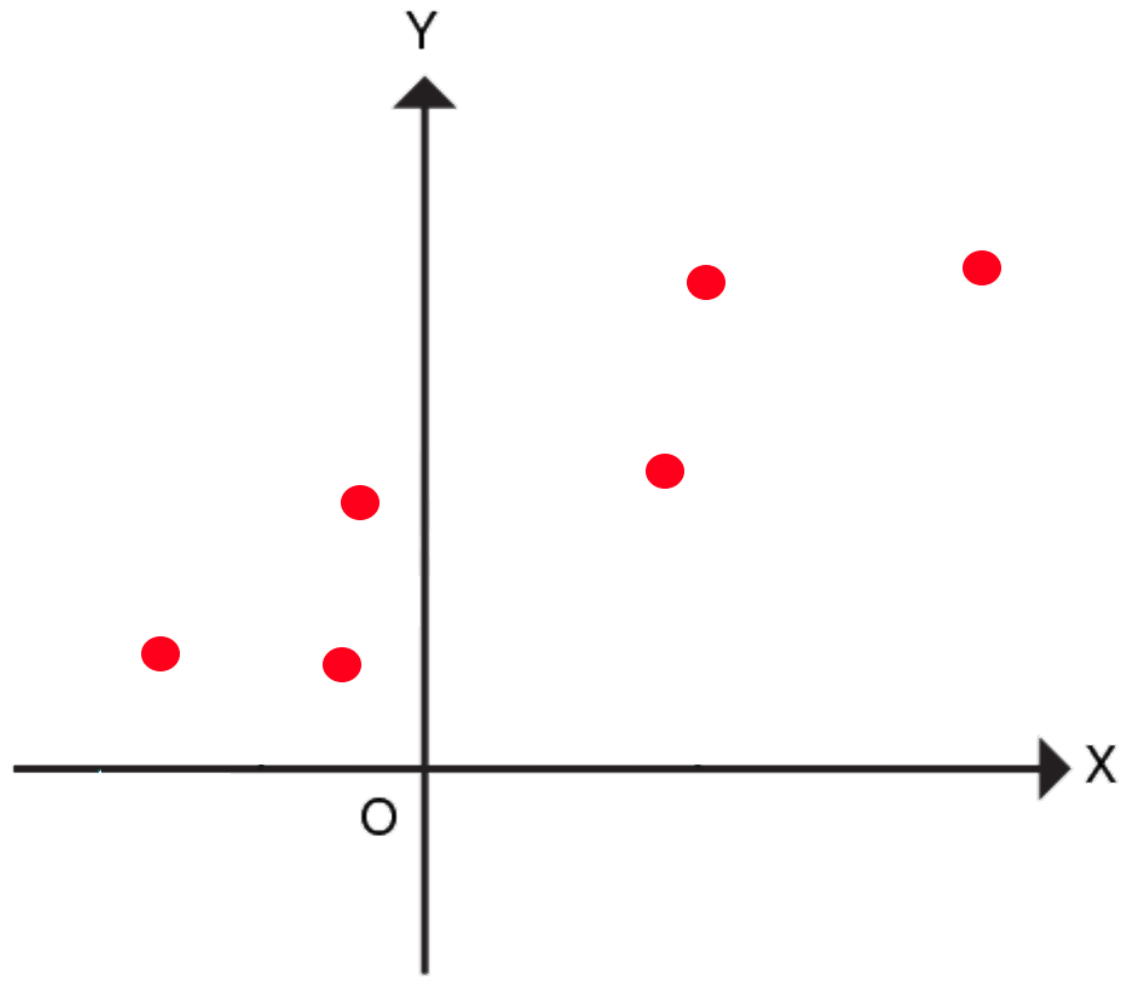}
   \caption{Raw values.}
\end{subfigure}
\begin{subfigure}[b]{0.45\linewidth}
   \includegraphics[width=\linewidth]{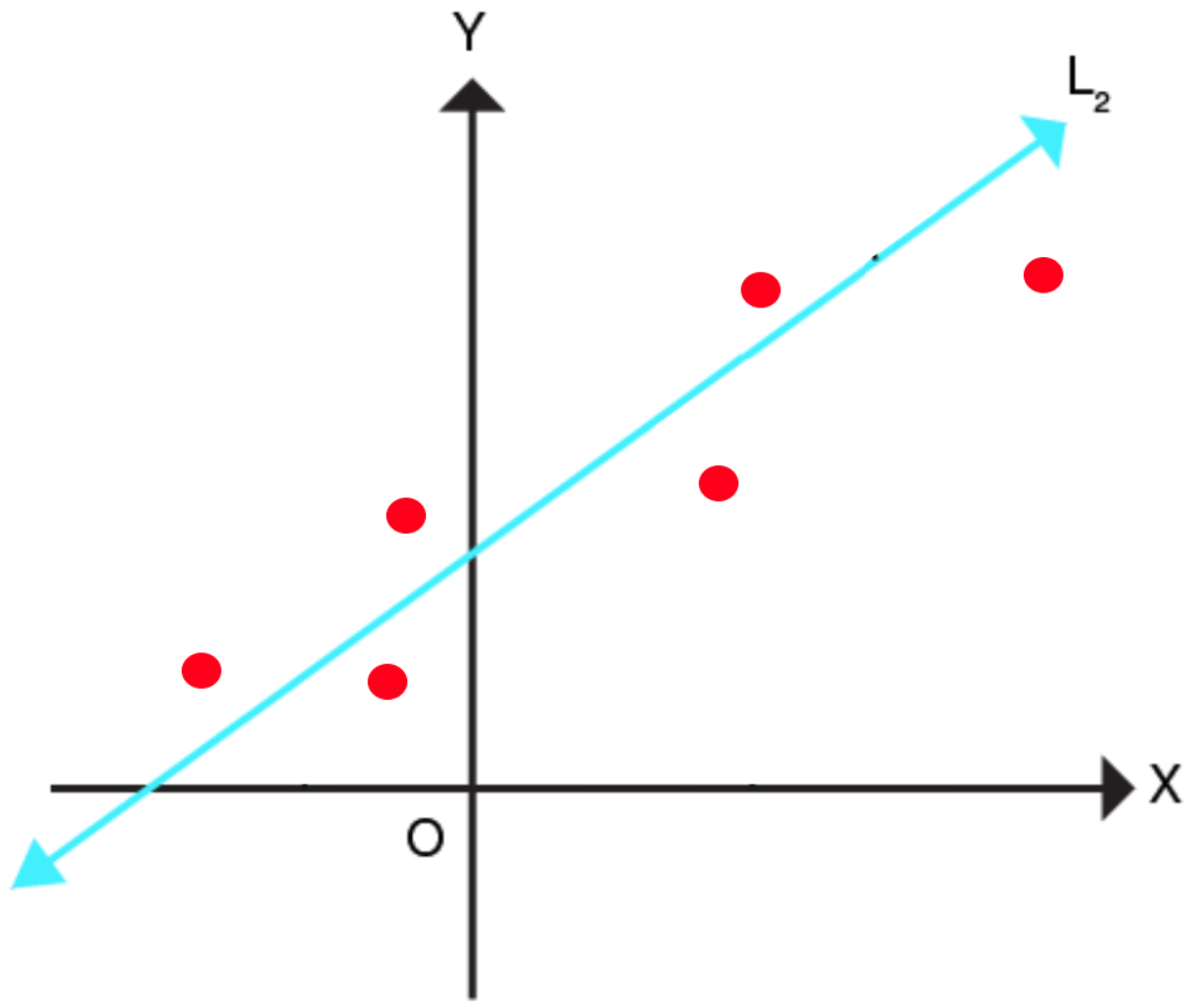}
   \caption{Linear regression.}
\end{subfigure}
\caption{\label{img:slope}Slope calculation of linear regression of the last values (window size of 6) of a meta-parameter.}
\end{figure}

During the training process, we calculate the progress of macroeconomic factors, commodity and index over the last few months (window size as meta-parameter) to analyze their trends. Specifically, we calculate the trend (slope of linear regression) direction (whether ascending or descending) of each macroeconomic factor, commodity, and index by examining their last monthly values over the specified window size (Figure~\ref{img:slope})\footnote{The slope calculation was done by the library scipy (\url{https://scipy.org/}).}. These calculated slopes for macroeconomic factors, commodities, and indices serve as the input data for our machine learning models, enabling predictions of business cycle phases.

\subsection{Evaluation metric}

All approaches were evaluated based on the accuracy of the Top1 and Top2 most probable phases of the business cycle\footnote{Total probability of falling into the first \textit{N} most probable regimes.}. Accuracy determines how close the candidates’ predictions are to their true labels:

\begin{equation}
    accuracy = \frac{1}{n_{samples}}\sum_{i=1}^{n_{samples}}1(\hat{y}_i = y_i)
\end{equation}

\noindent where $\hat{y}_i$ is the predicted value of the \textit{i-th} sample and $y_i$ is the corresponding true value.

To assess the effectiveness of the various approaches across different business phases, we conducted an analysis by calculating the F-score for each phase individually. Additionally, we computed the macro and weighted F-scores to offer a holistic evaluation encompassing all phases.

\subsection{Rule-based business cycle predictor}

In order to have a point of comparison, we created a method based on rules to define the phases of the business cycle from the growth and inflation indices described in Section~\ref{sc:dataset}. Inflation and growth are particularly important for gauging the current stage of an economic cycle. Equipped with the growth and inflation series, we can estimate the current phase within an economic cycle. To put this concept into practice, we built a method, namely as Rule-Based Business Cycle Predictor (RBBCP), based on the inflation and growth indices. This method analyzes the trends of these indices to estimate the phase of the business cycle. Table~\ref{tb:rule-based} summarizes the relation between the inflation and growth trends and their respective phases for the RBBCP method. For example, if inflation and growth have consistently decreased over the past few months, the method would predict the recession phase.

\begin{table}[h]
\centering
\begin{tabular}{ccc}
\hline
\textbf{Inflation} & \textbf{Growth} & \textbf{Phase} \\ \hline
Down               & Down            & Recession      \\
Down               & Up              & Recovery       \\
Up                 & Down            & Slowdown       \\
Up                 & Up              & Expansion      \\ \hline
\end{tabular}
\caption{\label{tb:rule-based}Rule-based business cycle predictor approach output based on the trends of inflation and growth indices.}
\end{table}

\subsection{Training procedure}

To set up the meta-parameters for each approach, we used the validation dataset. The window size was set to 12 and 9 for the EuroZone and the United States, respectively. Additionally, we have chosen not to include commodity and index data in the training process. In fact, these data did not display any or low correlation with the business cycle task. 

The MLP method is composed of four layers, each having 50 neurons. The last layer incorporates a dropout of 0.2 to improve the model's generalization ability. The training process of this model uses the Adam optimization algorithm with a learning rate of $0.005$ and the cross-entropy loss function.

Once the meta-parameters were defined, we proceeded to train the model using both the train and validation datasets.

\section{Experimental evaluation}
\label{sc:evaluation}

Tables~\ref{tb:ea-results} and~\ref{tb:us-results} summarize the performance of ML approaches on the test dataset for the EuroZone and United States, respectively.
Taking in consideration the business phases individually, MLR achieved the highest f-score values for all phases in US data. Meanwhile, in EuroZone, MLR outperformed other approaches only for the expansion and slowdown phases for EuroZone. MLP yielded better results for the recovery and recession phases in the EuroZone. 

The transitional phases (slowdown and recovery) present greater complexity for prediction compared to the relatively longer and more stable expansion and recession phases. Transition phases are distinguished by fluctuating trends and less clear-cut patterns, contributing to their challenging predictability.

\begin{table*}[h]
\scalebox{0.9}{
\begin{tabular}{lcccccccc}
\hline
\textbf{Approach} & \textbf{expansion} & \textbf{recovery} & \textbf{recession} & \textbf{slowdown} & \textbf{f-macro} & \textbf{f-weighted} & \textbf{Acc. T1} & \textbf{Acc. T2} \\
\hline
RBBCP    & 64.58\% & 38.3\% & 57.57\% & 35.29\% & 48.94\% & 53.1\% & 52.31\% & -- \\
MLR      & \textbf{75.55\%} & 53.33\% & 66.66\% & \textbf{53.66\%} & \textbf{62.30\%} & \textbf{65.78\%} & \textbf{65.25\%} & \textbf{84.74\%} \\
SVM & 74.16\% & 52.17\% & 64.40\% & 52.38\% & 60.78\% & 64.20\%& 63.56\%& 83.89\% \\
MLP & 73.68\%  & \textbf{55.81\%} & \textbf{68.85\%}  & 43.24\% & 60.40\% & 64.19\% & 64.41\% & 83.05\% \\
\hline
\end{tabular}
}
\caption{\label{tb:ea-results}Results for the EuroZone: F-score values categorized by economic regime, including macro and weighted F-scores, alongside Top1 and Top2 accuracies. The best results for each metric are in bold.}
\end{table*}

\begin{table*}[h]
\scalebox{0.9}{
\begin{tabular}{lcccccccc}
\hline
\textbf{Approach} & \textbf{expansion} & \textbf{recovery} & \textbf{recession} & \textbf{slowdown} & \textbf{f-macro} & \textbf{f-weighted} & \textbf{Acc. T1} & \textbf{Acc. T2} \\
\hline
RBBCP    & 63.15\% & 41.79\% & 45.45\% & 0\% & 37.6\% & 51.25\% & 47.65\% & -- \\
MLR      & \textbf{85.36\%} & \textbf{43.90\%} & \textbf{76.92\%} & \textbf{52.17\%} & \textbf{64.59\%} & \textbf{72.95\%} & \textbf{75\%}    & \textbf{92.14\%} \\
SVM      & 84.02\% & 39.13\% & 76\%    & 0\%     & 49.79\% & 68.05\% & 70.71\% & 87.86\% \\
MLP      & 83.83\% & 27.03\% & 62.50\% & 28.57\% & 50.48\% & 64.39\% & 67.14\% & 88.57\% \\
\hline
\end{tabular}
}
\caption{\label{tb:us-results}Results for the United States: F-score values categorized by economic regime, including macro and weighted F-scores, alongside Top1 and Top2 accuracies. The best results for each metric are in bold.}
\end{table*}


Considering all phases collectively, MLR enhanced the f-score values by achieving 62.3\% (macro) and 65.78\% (weighted) for the EuroZone, and by 64.59\% (macro) and 72.95\% (weighted) for the United States. Similarly, in both regions, MLR approach outperformed our baseline (RBBCP) and more complex approaches, such as SVM and MLP, in terms of Top1 and Top2 accuracies. 

As expected, RBBCP's analysis of the economic situation of a region is limited, which consequently affects its accuracy. The relatively poor performance of MLP can be attributed to the small size of the dataset. Deep learning approaches typically require large amounts of data to effectively train their models, which unfortunately, our monthly data of less than 500 examples does not provide. On the other hand, MLR and SVM are able to fit their models with less training data, which proved to be advantageous in this particular task.

Going deeper in the analysis of the MLR results, we examined the confusion matrix to identify instances where the predicted phase did not match the correct phase, and to determine which specific phases were responsible for these errors. 

Table~\ref{tb:confusion-matrix-us} shows the predictions for the United States. Most of the misclassified phases are between the "recovery and expansion" and "slowdown and recession" phases. These two pairs of phases can have similar characteristics in the economy, which makes it difficult to define the borderline between them. For instance, the recovery and expansion phases represent an improving economy where macroeconomic factors are getting better. Similarly, the slowdown and recession phases have their own particularities. This misclassification is evident in the accuracy of the Top2 most probable phases, where MLR achieved 92.14\% for the US. We found the level of accuracy, specifically, Top2 encouraging as in reality, economic regimes are not hardly discrete in particular when the economy is in transition. Thus, we reckon the ability of the model to provide probability for each regime appropriate and relevant.

\begin{table}[h]
\begin{adjustbox}{width=\linewidth}
\begin{tabular}{lcccc}
\hline
\backslashbox{Pred.}{Labels} & \textbf{recov.} & \textbf{expans.} & \textbf{slowd.} & \textbf{reces.} \\ \hline
recovery  &     9    &     19    &    0     &     2     \\ 
expansion &     2    &     70    &    2     &     1     \\ 
slowdown  &     0    &     0     &    6     &     2     \\ 
recession &     0    &     0     &    7     &     20    \\ \hline
\end{tabular}
\end{adjustbox}
\caption{\label{tb:confusion-matrix-us}Confusion matrix for the MLR approach on the test dataset for the United States.}
\end{table}

In the EuroZone, identifying borderlines between phases is more complicated than in the United States, and unfortunately, the model misclassified some cases between almost all phases (Table~\ref{tb:confusion-matrix-ea}). One of the most probable reasons for this is the complexity of analyzing the impact of macroeconomic factors of individual countries and how they collectively affect the economy of the EuroZone as a whole. The EuroZone is composed of several countries, which makes it more difficult to process and analyze their impact on the economy. 
Furthermore, it is important to note that the training data from 1981 to 2001 covers a period prior to the establishment of the Monetary Union, rendering the task of classifying the business cycle more intricate for a diverse and varied group of economies. During this period, economic data was likely computed and reported using disparate methodologies and protocols. Furthermore, it is essential to consider that the EuroZone initially consisted of twelve member states in 2002, and subsequently, eight additional nations joined between 2007 and 2023. Hence, the model's performance within this region is inevitably influenced by the multifaceted nature of this evolving economic landscape.

\begin{table}[h]
\begin{adjustbox}{width=\linewidth}
\begin{tabular}{lcccc}
\hline
\backslashbox{Pred.}{Labels} & \textbf{recov.} & \textbf{expans.} & \textbf{slowd.} & \textbf{reces.} \\ \hline
recovery  &     12   &    2      &     0    &     4     \\ 
expansion &     8    &    34     &     4    &     0     \\ 
slowdown  &     2    &    5      &     11   &     3     \\ 
recession &     5    &    3      &     5    &     20    \\ \hline
\end{tabular}
\end{adjustbox}
\caption{\label{tb:confusion-matrix-ea}Confusion matrix for the MLR approach on the test dataset for the EuroZone.}
\end{table}

Finally, we also analyzed the performance of these models taking in consideration only two labels: downswing (combination of recession and slowdown) and upswing (combination of expansion and recovery). Notably, we observed a significant improvement in the accuracy, resulting in 96.43\% and 80.51\% for the US and EZ, respectively. These results confirm the robustness of our methods for the business cycle task, whether using two or four labels.

To conclude, the business cycle task is a complex one, but the MLR approach has proven to be more effective than other machine learning approaches. In fact, the Top2 accuracy results showcase the strong performance of this approach, achieving 84.74\% accuracy for the EuroZone and 92.14\% accuracy for the United States. 

Despite the remarkable findings we have achieved, it is crucial to acknowledge that external factors and unforeseeable global events possess the capacity to significantly influence the global economy and precipitously alter the trajectories of businesses. Notably, the unforeseen emergence of the COVID-19 pandemic in 2019 and the ongoing conflict in Ukraine serve as salient illustrations of unforeseen events that can exert profound and immediate effects on the global economy, irrespective of economic performance and macroeconomic indicators.

\section{Conclusion}
\label{sc:conc}

Understanding the business cycle is crucial for promoting economic stability, guiding business planning, and informing investment decisions. The business cycle refers to the recurring pattern of macroeconomic factors that recognize the expansion and contraction of economic activity over time. These factors can include gross domestic product, employment rates, inflation, and consumer spending, among others. By monitoring the business cycle, policymakers, businesses, and investors can make more informed decisions, anticipate economic shifts, and prepare for potential risks and opportunities.

In this study, we compared three different machine learning approaches to forecast the phases of the business cycle, and the Multinomial Logistic Regression (MLR) achieved the best results. Specifically, MLR achieved an accuracy of 65.25\% (Top1) and 84.74\% (Top2) for the EuroZone and 75\% (Top1) and 92.14\% (Top2) for the United States. Many of the misclassified phases had similar characteristics and a blurred borderline between the transition of phases, making classification challenging. Despite these challenges, MLR approach demonstrated solid performance, as evidenced by the high accuracy achieved for the Top2 most probable phases.

Further work is currently underway to better analyze regions composed of several countries (e.g., EuroZone). We want to investigate how different macroeconomic factors for each country should be taken into account in the analysis of the economy as a whole area. We also intend to analyze and test different metrics and evaluation methods to better define the borderline of phases in the business cycle. Ultimately, we hope to set specific rules to differentiate similar phases and better define the transitions between these phases.

\section{Acknowledgements}

This work has been partially supported by the France Relance project (grant agreement number ANR-21-PRRD-0010-01).

\nocite{*}
\section{Bibliographical References}

\bibliographystyle{lrec-coling2024-natbib}
\bibliography{references}

\end{document}